# Bennett-type Generalization Bounds: Large-deviation Case and Faster Rate of Convergence

**Chao Zhang**
The Biodesign Institute
Arizona State University
Tempe, AZ 85287, USA

## Abstract

In this paper, we present the Bennett-type generalization bounds of the learning process for i.i.d. samples, and then show that the generalization bounds have a faster rate of convergence than the traditional results. In particular, we first develop two types of Bennett-type deviation inequality for the i.i.d. learning process: one provides the generalization bounds based on the uniform entropy number; the other leads to the bounds based on the Rademacher complexity. We then adopt a new method to obtain the alternative expressions of the Bennett-type generalization bounds, which imply that the bounds have a faster rate $o(N^{-\frac{1}{2}})$ of convergence than the traditional results $O(N^{-\frac{1}{2}})$. Additionally, we find that the rate of the bounds will become faster in the large-deviation case, which refers to a situation where the empirical risk is far away from (at least not close to) the expected risk. Finally, we analyze the asymptotical convergence of the learning process and compare our analysis with the existing results.

## 1 Introduction

In learning theory, one of the major concerns is to obtain the generalization bounds of the learning process for i.i.d. samples, which measure the probability that a function obtained in the i.i.d. learning process has a sufficiently small error [Vapnik, 1998, Bousquet et al., 2004]. Generalization bounds have been widely used to study many topics of learning theory, *e.g.*, the consistency of ERM-based learning processes [Vapnik, 1998], the asymptotic convergence of empirical process [Van der Vaart and Wellner, 1996] and the learnability of learning models [Blumer et al., 1989].

Deviation (or concentration) inequalities play an essential role in obtaining generalization bounds. There are many popular deviation and concentration inequalities, *e.g.*, Hoeffding's inequality, McDiarmid's inequality, Bennett's inequality, Bernstein's inequality and Talagrand's inequality. Among them, Hoeffding's inequality and McDiarmid's inequality have been intensively used to obtain the generalization bounds based on the covering number (or uniform entropy number) and the Rademacher complexity, respectively [Mendelson, 2003, Van der Vaart and Wellner, 1996]. To obtain the generalization bounds based on the VC dimension, Vapnik [1998] applied some classical inequalities, for example, Chernoff's inequality and Hoeffding's inequality, but also developed specific concentration inequalities. Bartlett et al. [2005] presented generalization bounds based on the local Rademacher complexity by using Talagrand's inequality. Additionally, there are also other works to study the generalization bound in statistical learning theory [Ben-David et al., 2010, Mohri and Rostamizadeh, 2010]. However, to our best knowledge, there is little theoretical investigation into the generalization bounds derived from Bennett's inequality.

### 1.1 Motivation

The paper is motivated by the difference between Hoeffding's inequality [Hoeffding, 1963] and Bennett's inequality [Bennett, 1962]. It is well-known that Hoeffding's inequality is achieved by only exploiting the expectation information, and Bennett's inequality is based on both of expectation and variance. Intuitively, the Bennett-type generalization bounds should have a faster rate of convergence than that of the Hoeffding-type results. However, to our best knowledge, this issue has not been well explored in the literature. In this paper, we will give a theoretical argument to show the faster rate of the Bennett-type results.

## 1.2 Overview of Main Results

In this paper, we are mainly concerned with the following three aspects: (1) Bennett-type deviation inequality; (2) Bennett-type generalization bounds; (3) rate of convergence.

By extending the classical Bennett's inequality, we use the martingale method to develop a Bennett-type deviation inequality for the case of multiple random variables and this inequality leads to the generalization bounds based on the uniform entropy number (UEN). Moreover, under the bounded-difference condition, we present another Bennett-type deviation inequality that is similar to McDiarmid's inequality. We use this deviation inequality to further obtain the generalization bounds based on the Rademacher complexity. Note that the aforementioned generalization bounds are said to be of Bennett-type, because they are derived from the Bennett-type deviation inequalities.

In order to analyze the rate of convergence of the bounds, one needs to obtain the alternative expressions of the Bennett-type generalization bounds. Differing from the Hoeffding-type bounds [see (7)], the expression of Bennett-type bounds [see (16)] are not well-defined and thus it is difficult to directly present the alternative expressions of the Bennett-type bounds.[1] Instead, one generally transforms the Bennett-type bound to the Bernstain-type one and then obtains the corresponding alternative expression showing that the bound has the rate $O(N^{-\frac{1}{2}})$ of convergence, which is in accordance with the classical result (8). Here, we use a new method to obtain another alternative expression of the Bennett-type bound and show that the Bennett-type bounds have a faster rate $o(N^{-\frac{1}{2}})$ of convergence than the classical result $O(N^{-\frac{1}{2}})$. Additionally, we point out that the rate of the bounds will become faster as the discrepancy between the empirical risk and the expected risk becomes larger. This situation is called as the large-deviation case [see Remark 4.3].

Note that it is well-known that the rate of the empirical Rademacher complexity is up to $O(N^{-\frac{1}{2}})$ [see (10)], and thus it seems to be a contradiction that the Bennett-type bounds have a faster rate $o(N^{-\frac{1}{2}}))$ [see Remark 5.1]. An explanation to the contradiction is given to support the theoretical findings on the faster rate of convergence.

## 1.3 Organization of the Paper

The rest of this paper is organized as follows. Section 2 formalizes the main issues of this paper and briefs the classical results. In Section 3, we present two Bennett-type deviation inequalities for the case of multiple random variables. Section 4 provides the generalization bounds based on the uniform entropy number (UEN). Moreover, under the bounded-difference condition, we present another Bennett-type deviation inequality and a new upper bound of the Rademacher complexity is presented in Section 5. In Section 6, we analyze the asymptotic convergence of the i.i.d. learning process and the last section concludes the paper. The proofs of the main results are given in the long version of this paper.[2]

## 2 Preliminaries

In this section, we first introduce some notations to formalize the proposed research of this paper, and then present the definitions of the covering number, the uniform entropy number and the Rademacher complexity. Moreover, we also brief the classical results of the Hoeffding-type generalization bounds of the i.i.d. learning process.

### 2.1 Problem Setup

Let $\mathcal{X} \subset \mathbb{R}^I$ and $\mathcal{Y} \subset \mathbb{R}^J$ be an input space and its corresponding output space, respectively. Denote $\mathcal{Z} := \mathcal{X} \times \mathcal{Y} \subset \mathbb{R}^I \times \mathbb{R}^J$ with $K = I + J$. Let $\mathcal{G} \subset \mathcal{Y}^{\mathcal{X}}$ be a function class with the domain $\mathcal{X}$ and the range $\mathcal{Y}$. Given a loss function $\ell : \mathcal{Y}^2 \to \mathbb{R}$, it is expected to find a function $g^* \in \mathcal{G} : \mathcal{X} \to \mathcal{Y}$ that minimizes the expected risk over $\mathcal{G}$

$$\mathrm{E}(\ell \circ g) := \int \ell(g(\mathbf{x}), \mathbf{y}) d\mathrm{P}(\mathbf{z}), \quad g \in \mathcal{G}, \qquad (1)$$

where $\mathrm{P}(\mathbf{z})$ stands for the distribution of $\mathbf{z} = (\mathbf{x}, \mathbf{y})$.

Generally, the distribution $\mathrm{P}(\mathbf{z})$ is unknown and thus the target function $g^*$ cannot be directly obtained by minimizing (1). Instead, we can apply the empirical risk minimization (ERM) principle to handle this issue [Vapnik, 1998]. Given a function class $\mathcal{G}$ and a set of i.i.d. samples $\mathbf{Z}_1^N := \{\mathbf{z}_n\}_{n=1}^N$ drawn from $\mathcal{Z}$, we define the empirical risk of $g \in \mathcal{G}$ as

$$\mathrm{E}_N(\ell \circ g) := \frac{1}{N} \sum_{n=1}^N \ell(g(\mathbf{x}_n), \mathbf{y}_n), \qquad (2)$$

which is considered as an approximation to the expected risk (1). Let $g_N \in \mathcal{G}$ be the function that minimizes the empirical risk (2) over $\mathcal{G}$ and deem $g_N$ as an estimate to $g^*$ with respect to $\mathbf{Z}_1^N$.

In the aforementioned i.i.d. learning process, we are mainly interested in the following two types of quantities:

---

[1] The reason is that it is difficult to directly obtain the analytical expression of the inverse function of $\Gamma(x) = x - (x+1)\ln(x+1)$.

[2] https://sites.google.com/site/czhang1015/

- $\mathrm{E}(\ell \circ g_N) - \mathrm{E}_N(\ell \circ g_N)$, which corresponds to the estimation of the expected risk from an empirical quantity;
- $\mathrm{E}(\ell \circ g_N) - \mathrm{E}(\ell \circ g^*)$, which corresponds to the performance of the ERM-based algorithm.

Recalling (1) and (2), since
$$\mathrm{E}_N(\ell \circ g^*) - \mathrm{E}_N(\ell \circ g_N) \geq 0,$$

we have
$$\begin{aligned}\mathrm{E}(\ell \circ g_N) =& \mathrm{E}(\ell \circ g_N) - \mathrm{E}(\ell \circ g^*) + \mathrm{E}(\ell \circ g^*) \\ \leq& \mathrm{E}_N(\ell \circ g^*) - \mathrm{E}_N(\ell \circ g_N) + \mathrm{E}(\ell \circ g_N) \\ & - \mathrm{E}(\ell \circ g^*) + \mathrm{E}(\ell \circ g^*) \\ \leq& 2 \sup_{g \in \mathcal{G}} \big|\mathrm{E}(\ell \circ g) - \mathrm{E}_N(\ell \circ g)\big| + \mathrm{E}(\ell \circ g^*),\end{aligned}$$

and thus
$$0 \leq \mathrm{E}(\ell \circ g_N) - \mathrm{E}(\ell \circ g^*) \leq 2 \sup_{g \in \mathcal{G}} \big|\mathrm{E}(\ell \circ g) - \mathrm{E}_N(\ell \circ g)\big|,$$

with
$$\mathrm{E}(\ell \circ g_N) - \mathrm{E}_N(\ell \circ g_N) \leq \sup_{g \in \mathcal{G}} \big|\mathrm{E}(\ell \circ g) - \mathrm{E}_N(\ell \circ g)\big|.$$

This shows that the asymptotic behaviors of the aforementioned two quantities, when the sample number $N$ goes to *infinity*, can both be described by the supremum
$$\sup_{g \in \mathcal{G}} \big|\mathrm{E}(\ell \circ g) - \mathrm{E}_N(\ell \circ g)\big|, \qquad (3)$$
which is the so-called generalization bound of the i.i.d. learning process.

For convenience, we define the loss function class
$$\mathcal{F} := \{\mathbf{z} \mapsto \ell(g(\mathbf{x}), \mathbf{y}) : g \in \mathcal{G}\},$$
and call $\mathcal{F}$ as the function class in the rest of this paper. By (1) and (2), given a sample set $\mathbf{Z}_1^N$ drawn from $\mathcal{Z}$, we briefly denote for any $f \in \mathcal{F}$,
$$\mathrm{E}f := \int f(\mathbf{z}) d\mathrm{P}(\mathbf{z}); \; \mathrm{E}_N f := \frac{1}{N} \sum_{n=1}^N f(\mathbf{z}_n).$$

Thus, we rewrite the generalization bound (3) as
$$\sup_{f \in \mathcal{F}} \big|\mathrm{E}f - \mathrm{E}_N f\big|.$$

## 2.2 Complexity Measures of Function Classes

Generally, the generalization bound of a certain learning process is achieved by incorporating the complexity measure of the function class, *e.g.*, the covering number, the VC dimensions and the Rademacher complexity. This paper is mainly concerned with the covering number, the uniform entropy number (UEN) and the Rademacher complexity.

### 2.2.1 Covering Number and Uniform Entropy Number (UEN)

The following is the definition of the covering number and we refer to Mendelson [2003] for details.

**Definition 2.1** *Let $\mathcal{F}$ be a function class and $d$ be a metric on $\mathcal{F}$. For any $\xi > 0$, the covering number of $\mathcal{F}$ at radius $\xi$ with respect to the metric $d$, denoted by $\mathcal{N}(\mathcal{F}, \xi, d)$ is the minimum size of a cover of radius $\xi$.*

For clarity of presentation, we give a useful notation for the following discussion. Given a sample set $\mathbf{Z}_1^N := \{\mathbf{z}_n\}_{n=1}^N$ drawn from $\mathcal{Z}$, we denote $\mathbf{Z'}_1^N := \{\mathbf{z'}_n\}_{n=1}^N$ as the ghost-sample set drawn from $\mathcal{Z}$ such that the ghost sample $\mathbf{z'}_n$ has the same distribution as $\mathbf{z}_n$ for any $1 \leq n \leq N$. Denote $\mathbf{Z}_1^{2N} := \{\mathbf{Z}_1^N, \mathbf{Z'}_1^N\}$. Setting the metric $d$ as the $\ell_p(\mathbf{Z}_1^{2N})$ ($p > 0$) norm, we then obtain the covering number $\mathcal{N}\big(\mathcal{F}, \xi, \ell_p(\mathbf{Z}_1^{2N})\big)$.

The uniform entropy number (UEN) is a variant of the covering number and we refer to Mendelson [2003] for details as well. By setting the metric $\ell_p(\mathbf{Z}_1^N)$ ($p > 0$), the UEN is defined as follows:
$$\ln \mathcal{N}_p(\mathcal{F}, \xi, N) := \sup_{\mathbf{Z}_1^{2N}} \ln \mathcal{N}\big(\mathcal{F}, \xi, \ell_p(\mathbf{Z}_1^N)\big). \quad (4)$$

### 2.2.2 Rademacher Complexity

The Rademacher complexity is one of the most frequently used complexity measures of function classes and we refer to Bousquet et al. [2004] for details.

**Definition 2.2** *Let $\mathcal{F}$ be a function class and $\{\mathbf{z}_n\}_{n=1}^N$ be a sample set drawn from $\mathcal{Z}$. Denote $\{\sigma_n\}_{n=1}^N$ as a set of random variables independently taking either value of $\{-1, 1\}$ with equal probability. The Rademacher complexity of $\mathcal{F}$ is defined as*
$$\mathcal{R}(\mathcal{F}) := \mathrm{E} \sup_{f \in \mathcal{F}} \left\{ \frac{1}{N} \sum_{n=1}^N \sigma_n f(\mathbf{z}_n) \right\} \quad (5)$$

*with its empirical version*
$$\mathcal{R}_N(\mathcal{F}) := \mathrm{E}_\sigma \sup_{f \in \mathcal{F}} \left\{ \frac{1}{N} \sum_{n=1}^N \sigma_n f(\mathbf{z}_n) \right\}, \quad (6)$$

*where $\mathrm{E}$ stands for the expectation taken with respect to all random variables $\{\mathbf{z}_n\}_{n=1}^N$ and $\{\sigma_n\}_{n=1}^N$, and $\mathrm{E}_\sigma$ stands for the expectation only taken with respect to random variables $\{\sigma_n\}_{n=1}^N$.*

## 2.3 Classical Hoeffding-type Generalization Bounds

Next, we summarize some classical results of the generalization bounds. Note that the following bounds are said to be of Hoeffding-type, because they are derived from (or strongly related to) Hoeffding's inequality and Hoeffding's lemma [Hoeffding, 1963].

First, based on the covering number, one can use Hoeffding's inequality (or Hoeffding's lemma) to obtain the following generalization bound w.r.t. a function class $\mathcal{F}$ with the range $[a, b]$ [see Mendelson, 2003, Theorem 2.3]: for any $N \geq \frac{8(b-a)^2}{\xi^2}$,

$$\Pr \left\{ \sup_{f \in \mathcal{F}} \left| \mathrm{E}f - \mathrm{E}_N f \right| > \xi \right\}$$
$$\leq 8 \mathrm{E} \mathcal{N}\left(\mathcal{F}, \xi/8, \ell_1(\mathbf{Z}_1^{2N})\right) \exp \left\{ -\frac{N\xi^2}{32(b-a)^2} \right\}, \quad (7)$$

which is one of the most frequently used generalization results in statistical learning theory. Because of its well-defined expression, we can directly obtain its alternative expression: for any $N \geq \frac{8(b-a)^2}{\xi^2}$, with probability at least $1 - \epsilon$,

$$\sup_{f \in \mathcal{F}} \left| \mathrm{E}_N f - \mathrm{E}f \right| \qquad (8)$$
$$\leq O\left( \left( \frac{\ln \mathrm{E} \mathcal{N}\left(\mathcal{F}, \xi/8, \ell_1(\mathbf{Z}_1^{2N})\right) - \ln(\epsilon/8)}{N} \right)^{\frac{1}{2}} \right).$$

Following the alternative expression (8), it is observed that the generalization bound $\sup_{f \in \mathcal{F}} \left| \mathrm{E}_N f - \mathrm{E}f \right|$ has a convergence rate of $O(N^{-\frac{1}{2}})$.

On the other hand, McDiarmid's inequality can provide the following generalization bound based on the Rademacher complexity [see Bousquet et al., 2004, Theorem 5]: for any $\epsilon > 0$ and $f \in \mathcal{F}$, with probability at least $1 - \epsilon$,

$$\mathrm{E}f \leq \mathrm{E}_N f + 2\mathcal{R}(\mathcal{F}) + (b-a)\left(\frac{\ln(1/\epsilon)}{N}\right)^{\frac{1}{2}}$$
$$\leq \mathrm{E}_N f + 2\mathcal{R}_N(\mathcal{F}) + 3(b-a)\left(\frac{\ln(2/\epsilon)}{2N}\right)^{\frac{1}{2}}, \quad (9)$$

which is also of Hoeffding-type, because McDiarmid's inequality is actually derived from Hoeffding's lemma under the condition that $f$ has the bounded-differences property [see Bousquet et al., 2004, Theorem 6]. Furthermore, it is followed from Sudakov minoration for Rademacher processes[3] [Talagrand, 1994b,a, Latała, 1997] and Massart's finite class lemma (Dudley's entropy integral)[4] [Mendelson, 2003, Van der Vaart and

---
[3]See http://www.cs.berkeley.edu/ bartlett/courses/281b-sp08/19.pdf
[4]See http://ttic.uchicago.edu/∼karthik/dudley.pdf

Wellner, 1996] that:

$$\frac{c}{\ln N} \sup_{\alpha > 0} \alpha \sqrt{\frac{\ln \mathcal{N}(\mathcal{F}, \xi, \ell_2(\mathbf{Z}_1^N))}{N}} \leq \mathcal{R}_N(\mathcal{F}) \quad (10)$$
$$\leq \inf_{\epsilon > 0} \left\{ 4\epsilon + 12 \int_\epsilon^\infty \sqrt{\frac{\ln \mathcal{N}(\mathcal{F}, \xi, \ell_2(\mathbf{Z}_1^N))}{N}} d\xi \right\},$$

which shows the lower and the upper bounds of the empirical Rademacher complexity $\mathcal{R}_N(\mathcal{F})$. The combination of (9) and (10) implies that the rate of convergence of $\sup_{f \in \mathcal{F}} \left| \mathrm{E}_N f - \mathrm{E}f \right|$ is up to $O(N^{-\frac{1}{2}})$ as well.

## 3 Bennett-type Deviation Inequalities

By extending the classical Bennett's inequality [Bennett, 1962] to the case of multiple random variables, this section will present two Bennett-type deviation inequalities for the i.i.d. learning process. We also refer to Bousquet [2002] for the application of Bennett's inequality in the empirical process.

**Theorem 3.1** *Let $f$ be a function with the range $[a, b]$ and $\mathbf{Z}_1^N = \{\mathbf{z}_n\}_{n=1}^N$ be a set of i.i.d. samples drawn from $\mathcal{Z}$. Define a function $F : \mathbb{R}^{KN} \to \mathbb{R}$ as*

$$F\left(\mathbf{Z}_1^N\right) := \sum_{n=1}^N f(\mathbf{z}_n). \quad (11)$$

*Then, we have for any $0 < \xi < N(b - a)$,*

$$\Pr\left\{ \left| \mathrm{E}\{F\} - F(\mathbf{Z}_1^N) \right| > \xi \right\} \leq 2 \mathrm{e}^{N\Gamma\left(\frac{\xi}{N(b-a)}\right)}, \quad (12)$$

*where*

$$\Gamma(x) := x - (1 + x)\ln(1 + x). \quad (13)$$

The proof of this result is processed by the martingale method. Compared to the classical Bennett's inequality, this result is valid for the case of multiple random variables and provides the convenience to obtain the generalization bound of the i.i.d. learning process. Especially, the two inequalities will coincide, if there is only one random variable.

Moreover, recalling the classical McDiarmid's inequality [see Bousquet et al., 2004, Theorem 6], it is actually derived from Hoeffding's lemma under the condition that $f$ has the bounded-differences property. Thus, McDiarmid's inequality has a similar expression to that of Hoeffding's inequality and the two inequalities coincide if there is only one random variable. Similarly, we obtain another Bennett-type deviation inequality under the bounded-difference condition:

**Theorem 3.2** *Let $\mathbf{z}_1, \cdots, \mathbf{z}_N$ be $N$ independent random variables taking values from $\mathcal{Z}$. Assume that*

*there exists a positive constant such that the function $H : \mathcal{Z}^N \to \mathbb{R}$ satisfies the bounded-difference condition: for any $1 \leq n \leq N$,*

$$\sup_{\mathbf{z}_1,\cdots,\mathbf{z}_N,\mathbf{z}'_n} \Big| H(\mathbf{z}_1,\cdots,\mathbf{z}_n,\cdots,\mathbf{z}_N) \\ - H(\mathbf{z}_1,\cdots,\mathbf{z}'_n,\cdots,\mathbf{z}_N) \Big| \leq c. \quad (14)$$

*Then, we have for any $\xi > 0$*

$$\Pr\Big\{ H(\mathbf{z}_1,\cdots,\mathbf{z}_n,\cdots,\mathbf{z}_N) \qquad (15)\\ - \mathrm{E}H(\mathbf{z}_1,\cdots,\mathbf{z}_n,\cdots,\mathbf{z}_N) \geq \xi \Big\} \leq \exp\left\{N\Gamma\left(\frac{\xi}{Nc}\right)\right\},$$

*where $\Gamma(x)$ is defined in (13).*

The inequality (15) is an extension of the classical Bennett's inequality and the two results will coincide if there is only one random variable as well.

Subsequently, we will use the above two deviation inequalities to obtain the generalization bounds based on the uniform entropy number and the Rademacher complexity, respectively.

## 4 Bennett-type Generalization Bounds

In this section, we will show the Bennett-type generalization bounds of the i.i.d learning process. The derived bounds are based on the uniform entropy number (UEN) and the Rademacher complexity, respectively. Note that since the presented bounds (16), (18) and (23) are derived from the Bennett-type deviation inequalities, they are said to be of Bennett-type as well.

### 4.1 UEN-based Generalization Bounds

By using the deviation inequality (12) and the symmetrization inequality [see Bousquet et al., 2004, Lemma 2], we can obtain a Bennett-type generalization bound based on the uniform entropy number:

**Theorem 4.1** *Assume that $\mathcal{F}$ is a function class with the range $[a, b]$. Let $\mathbf{Z}_1^N$ and $\mathbf{Z}'^N_1$ be drawn from $\mathcal{Z}$ and denote $\mathbf{Z}_1^{2N} := \{\mathbf{Z}_1^N, \mathbf{Z}'^N_1\}$. Then, given any $0 < \xi \leq (b-a)$, we have for any $N \geq \frac{8(b-a)^2}{\xi^2}$,*

$$\Pr\left\{\sup_{f \in \mathcal{F}} |\mathrm{E}f - \mathrm{E}_N f| > \xi\right\} \qquad (16) \\ \leq 8\mathcal{N}_1(\mathcal{F}, \xi/8, 2N) \exp\left\{N\Gamma\left(\frac{\xi}{8(b-a)}\right)\right\}.$$

This theorem implies that the probability of the event that the generalization bound $\sup_{f \in \mathcal{F}} |\mathrm{E}f - \mathrm{E}_N f|$ is larger than any $\xi > 0$ can be bounded by the right-hand side of (16). We can find that the expression of the bound is similar to that of Bennett's inequality.

Different from the aforementioned Hoeffding-type result (7) and its alternative expression (8), it is difficult to directly achieve the alternative expression of the Bennett-type bound (16), because it is difficult to obtain the analytical expression of the inverse function of $\Gamma(x) = x - (x+1)\ln(x+1)$. Instead, one generally uses the term $\frac{-x^2}{2+(2x/3)}$ to approximate the function $\Gamma(x)$ and then get the so-called Bernstein's inequality.[5] In the same way, we can obtain the following alternative expression of the Bennett-type result (16):

$$\sup_{f \in \mathcal{F}} |\mathrm{E}_N f - \mathrm{E}f| \qquad (17) \\ \leq \frac{4(b-a)\big(\ln \mathcal{N}_1(\mathcal{F}, \xi/8, 2N) - \ln(\epsilon/8)\big)}{3N} \\ + \frac{(b-a)\sqrt{2\big(\ln \mathcal{N}_1(\mathcal{F}, \xi/8, 2N) - \ln(\epsilon/8)\big)}}{\sqrt{N}},$$

which implies that the rate of convergence of the Bennett-type bound is also up to $O(N^{-\frac{1}{2}})$. It is in accordance with the rate of the aforementioned classical results (8). For convenience, the alternative expression is said to be of Bernstein-type if no confusion arises.

### 4.2 Bennett-type Alternative Expression and Faster Rate of Convergence

Recalling the process of obtaining Hoeffding's inequality and Bennett's inequality, the former is achieved by using the information of expectation, while the latter needs to consider the information of both expectation and variance. Intuitively, the Bennett-type result (16) should have a faster rate of convergence than that of the Hoeffding-type result (7). From this point of view, we introduce a new method to obtain another alternative expression of the Bennett-type result (16) and show that the rate of the generalization bound $\sup_{f \in \mathcal{F}} |\mathrm{E}_N f - \mathrm{E}f|$ can reach $o(N^{-\frac{1}{2}})$, when $N$ goes to infinity:

**Theorem 4.2** *Follow the notations and conditions of Theorem 4.1. Then, given any $0 < \xi \leq (b-a)$ and for any $N \geq \frac{8(b-a)^2}{\xi^2}$, we have with probability at least $1 - \epsilon$,*

$$\sup_{f \in \mathcal{F}} |\mathrm{E}_N f - \mathrm{E}f| \leq \qquad (18) \\ 8(b-a)\left(\frac{\ln \mathcal{N}_1(\mathcal{F}, \xi/8, 2N) - \ln(\epsilon/8)}{\beta_1 N}\right)^{\frac{1}{\gamma}}.$$

---
[5]http://ocw.mit.edu/courses/mathematics/18-465-topics-in-statistics-statistical-learning-theory-spring-2007/lecture-notes/l6.pdf

where $\beta_1 \in (0.0075, 0.4804)$,

$$\epsilon := 8\mathcal{N}_1(\mathcal{F}, \xi/8, 2N) \exp\{N\Gamma(x)\},$$

and $0 < \gamma(\beta_1; x) \leq \gamma < 2$ ($x \in (0, 1/8]$) with

$$\gamma(\beta; x) := \frac{\ln\left(((x+1)\ln(x+1) - x)/\beta\right)}{\ln x}. \quad (19)$$

**Proof.** Since any $f \in \mathcal{F}$ is a bounded function with the range $[a, b]$, the supremum $\sup_{f \in \mathcal{F}} |\mathrm{E}f - \mathrm{E}_N f|$ will not be larger than $(b - a)$. Therefore, the quantity $\frac{\xi}{8(b-a)}$ in (16) will not be larger than $1/8$. Set $x = \frac{\xi}{8(b-a)}$ with $x \in (0, 1/8]$ and consider the following equation with respect to $\gamma > 0$

$$\Gamma(x) = x - (x+1)\ln(x+1) = -\beta_1(x^\gamma), \quad (20)$$

where $\beta_1$ is some positive constant. Denote the solution to the equation (20) w.r.t. $\gamma$ as

$$\gamma(\beta_1, x) := \frac{\ln\left(\frac{(x+1)\ln(x+1) - x}{\beta_1}\right)}{\ln(x)}. \quad (21)$$

By numerical simulation, we find that given any $\beta_1 \in (0.0075, 0.4804)$, there holds that $0 < \gamma(\beta_1; x) < 2$ for any $x \in (0, 1/8]$ (see Fig. 1). Then, given any $x \in (0, 1/8]$ and $\beta_1 \in (0.0075, 0.4804)$, we have for any $\widetilde{\gamma} \in [\gamma(\beta_1; x), 2)$,

$$x - (x+1)\ln(x+1) \leq -\beta_1 x^{\widetilde{\gamma}} < -\beta_1 x^2. \quad (22)$$

By combining Theorem 4.1 and (22), we can straightforwardly show an upper bound of the generalization bound $\sup_{f \in \mathcal{F}} |\mathrm{E}_N f - \mathrm{E}f|$: letting

$$\epsilon := 8\mathcal{N}_1(\mathcal{F}, \xi/8, 2N) \exp\{N\Gamma(x)\},$$

and with probability at least $1 - \epsilon$,

$$\sup_{f \in \mathcal{F}} |\mathrm{E}_N f - \mathrm{E}f|$$
$$\leq 8(b-a) \left(\frac{\ln \mathcal{N}_1(\mathcal{F}, \xi/8, 2N) - \ln(\epsilon/8)}{\beta_1 N}\right)^{\frac{1}{\gamma}}.$$

where $0 < \gamma(\beta_1; x) \leq \gamma < 2$ with $x \in (0, 1/8]$. ∎

The above theorem shows a Bennett-type alternative expression of the generalization bound $\sup_{f \in \mathcal{F}} |\mathrm{E}_N f - \mathrm{E}f|$, which provides a faster rate $o(N^{-\frac{1}{2}})$ of convergence than the rate $O(N^{-\frac{1}{2}})$ of the classical result (8) and the Bernstein-type result (17). The main starting point of this theorem is whether there exists a positive constant $\beta_1$ such that the function $\gamma(\beta_1; x)$ is smaller than 2 for any $x \in (0, 1/8]$ [see (20) and (21)], i.e., there holds that $0 < \gamma(\beta_1; x) < 2$ for any $0 < x \leq$

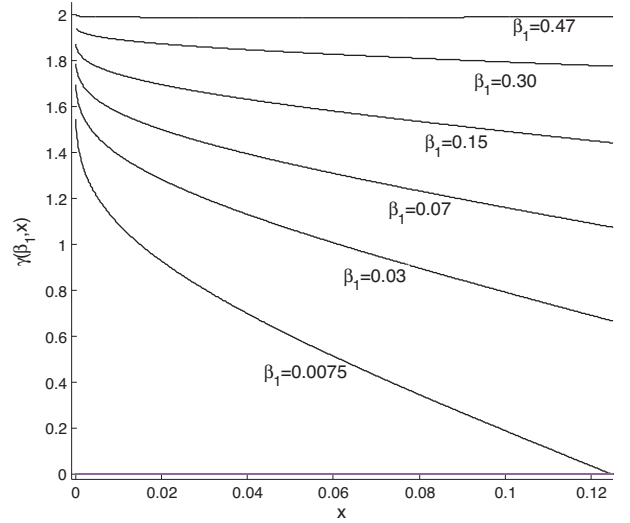

Figure 1: The Function Curve of $\gamma(\beta_1; x)$

$1/8$. In the proof, we show that the value interval $(0.0075, 0.4804)$ of $\beta_1$ supports $\Gamma(x) = -\beta_1 x^{\gamma(\beta_1; x)} < -\beta_1 x^2$ for any $x \in (0, 1/8]$. As shown Fig. 1, given any $\beta_1 \in (0.0075, 0.4804)$, the function $\gamma(\beta_1; x)$ is smaller than 2 w.r.t. $x \in (0, 1/8]$. Meanwhile, given any $x \in (0, 1/8]$, the function $\gamma(\beta_1; x)$ is monotonically increasing w.r.t. $\beta_1 \in (0.0075, 0.4804)$. However, the function $\gamma(\beta_1; x)$ is not monotonically decreasing w.r.t. $x \in (0, 1/8]$ for any $\beta_1 \in (0.0075, 0.4804)$. In fact, $\gamma(\beta_1; x)$ is monotonically decreasing when $\beta_1 \in (0.0075, 0.4434]$ and has one single minimizer $x_0 \in (0, 1/8]$ of $\gamma(\beta_1; x)$ when $\beta_1 \in (0.4434, 0.4804)$ with $\gamma'(\beta_1; x_0) = 0$.

### 4.3 Large-deviation Case

Subsequently, we will show that the Bennett-type bounds can reach a faster rate of convergence in the large-deviation case.

**Remark 4.3** *The word "large-deviation" means that the discrepancy between the empirical risk and the expected risk is large (or not small). Given any $\xi > 0$, one of our major concerns is the probability $\Pr\{\sup_{f \in \mathcal{F}} |\mathrm{E}_N f - \mathrm{E}f| > \xi\}$, and then we say that the case the value of $\xi$ approaches to $(b-a)$ is of large-deviation.*[6]

Actually, the large-deviation case is referring to a situation where the empirical quantity $\mathrm{E}_N f$ is far away from (at least not close to) the expected risk $\mathrm{E}f$. As shown in Fig. 1, for any $\beta_1 \in (0.0075, 0.4434]$, the curve of $\gamma(\beta_1; x)$ is monotonically decreasing w.r.t.

---

[6]The function class $\mathcal{F}$ is composed of bounded functions with the range $[a, b]$.

$x \in (0, 1/8]$, and for any $\beta_1 \in (0.4434, 0.4804)$ the function $\gamma(\beta_1; x)$ is still smaller than 2 for any $x \in (0, 1/8]$. Moreover, there holds that for any $\beta_1 \in (0.0075, 0.4804)$,

$$\lim_{x \to 0+} \gamma(\beta_1, x) := \lim_{x \to 0+} \frac{\ln\left(\frac{(x+1)\ln(x+1)-x}{\beta_1}\right)}{\ln(x)} = 2.$$

This illustrates that the rate $O(N^{-\frac{1}{\gamma}})$ ($\gamma(\beta_1; x) \leq \gamma$) of the Bennett-type generalization bound (18) becomes faster as the empirical quantity $\mathrm{E}_N f$ goes further away from the expected risk $\mathrm{E}f$ (i.e. $x$ approaches to $1/8$). However, when $\mathrm{E}_N f$ approaches to $\mathrm{E}f$ (i.e. $x$ goes to 0), the rate $O(N^{-\frac{1}{\gamma}})$ will approach to $O(N^{-\frac{1}{2}})$, which is in accordance with the classical Hoeffding-type results.

In contrast, the Hoeffding-type results consistently provide the rate $O(N^{-\frac{1}{2}})$ regardless of the discrepancy between $\mathrm{E}_N f$ and $\mathrm{E}f$. Thus, the Bennett-type results can give a more detailed description to the asymptotical behavior of the learning process.

### 4.4 Effect of the Parameter $\beta_1$

As addressed in the introduction, the main motivation of this paper is to study whether the Bennett-type generalization bounds have a faster rate of convergence than that of the Hoeffding-type generalization results, because the Bennett-type results are derived from the corporation of the expectation information and the variance information and in contrast, the Hoeffding-type results are only related to the expectation information. However, the main challenge to analyze the rate lies in obtaining the alternative expression of the bound (16), because it is difficult to analytically express the inverse function of $\Gamma(x)$. Instead, we exploit the term $-\beta x^\gamma$ to substitute $\Gamma(x)$ [see (20)] and Fig. 2 illustrates the validity of this method. In Fig. 2, the curve $\mathrm{e}^{-x^2/32}$ and $\mathrm{e}^{\Gamma(x/8)}$ correspond to the Hoeffding-type bound (7) and the Bennett-type bound (16), respectively. Evidently, setting $\beta_1 = 0.4804$ makes the curve $\mathrm{e}^{-\beta_1(x/8)^2}$ almost coincide with the curve $\mathrm{e}^{\Gamma(x/8)}$.

### 4.5 Rademacher-complexity-based Generalization Bounds

By using the deviation inequality (15), we can obtain the generalization bounds based on the Rademacher complexity:

**Theorem 4.4** *Assume that $\mathcal{F}$ is a function class consisting of functions with the range $[a, b]$. Let $\mathbf{Z}_1^N = \{\mathbf{z}_n\}_{n=1}^N$ be a set of i.i.d. samples drawn from $\mathcal{Z}$. Then, for any $f \in \mathcal{F}$, we have with probability at least*

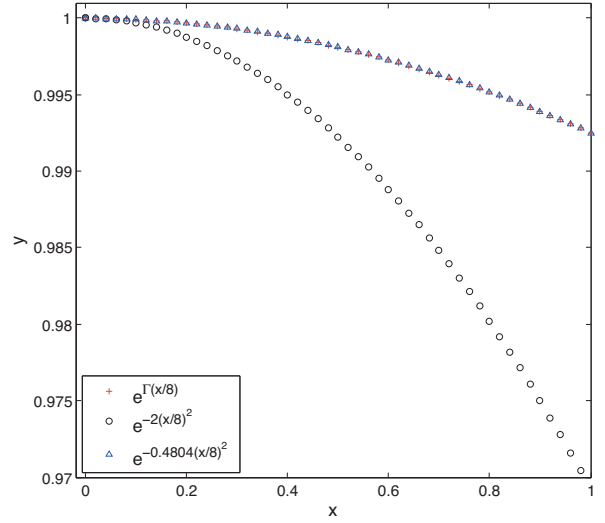

Figure 2: The Function Curves of $\mathrm{e}^{\Gamma(x/8)}$, $\mathrm{e}^{-x^2/32}$ and $\mathrm{e}^{-0.4804(x/8)^2}$.

$1 - \epsilon$,

$$\mathrm{E}f < \mathrm{E}_N f + 2\mathcal{R}(\mathcal{F}) + (b-a)\left(\frac{\ln(1/\epsilon)}{\beta_2 N}\right)^{\frac{1}{\gamma}}$$

$$< \mathrm{E}_N f + 2\mathcal{R}_N(\mathcal{F}) + 3(b-a)\left(\frac{\ln(2/\epsilon)}{\beta_2 N}\right)^{\frac{1}{\gamma}}, \quad (23)$$

*where $\beta_2$ is taken from the interval $(0.0075, 0.3863)$, $\epsilon := \exp\{N\Gamma(x)\}$ and $0 < \gamma(\beta_2; x) \leq \gamma < 2$ ($x \in (0, 1]$) with $\gamma(\beta; x)$ defined in (19).*

The above theorem presents the Bennett-type generalization bounds based on the Rademacher complexity. Compared to the classical Rademacher-complexity-based results (9), the new bounds (23) incorporate a term $O(N^{-\frac{1}{\gamma}})$ (i.e., $o(N^{-\frac{1}{2}})$ because $\gamma < 2$) with a faster rate of convergence than the term $O(N^{-\frac{1}{2}})$ appearing in (9), when $N$ goes to infinity.

## 5 Bounds of Rademacher Complexities

Recalling (16) and (23), it is observed that there might be a contradiction:

**Remark 5.1** *The generalization bound (16) has the faster rate $o(N^{-\frac{1}{2}})$, while the magnitude of the empirical Rademacher complexity has been proved to be $O(N^{-\frac{1}{2}})$ as shown in the classical result (10).*

In order to explain this contradiction, we should consider the following two questions:

**(Q1)** whether the empirical Rademacher complexity $\mathcal{R}_N(\mathcal{F})$ can totally describe the behavior of Rademacher complexity $\mathcal{R}(\mathcal{F})$;

**(Q2)** whether the classical results (10) are either applicable to $\mathcal{R}(\mathcal{F})$.

### 5.1 Answer to Question Q1

Recalling (9) and (23), the quantity $(\mathrm{E}f - \mathrm{E}_N f)$ is originally bounded by the Rademacher complexity $\mathcal{R}(\mathcal{F})$, while it is difficult to compute $\mathcal{R}(\mathcal{F})$ due to the unknown distribution of $\mathbf{Z}_1^N$. Instead, by using deviation inequalities (*e.g.* Hoeffding's inequality and Bennett's inequality), one can further bound $\mathcal{R}(\mathcal{F})$ by using its empirical version $\mathcal{R}_N(\mathcal{F})$ as follows: with probability as least $1 - \epsilon/2$,

$$\mathcal{R}(\mathcal{F}) \leq \mathcal{R}_N(\mathcal{F}) + (b-a)\left(\frac{\ln(2/\epsilon)}{2N}\right)^{\frac{1}{2}},$$

and

$$\mathcal{R}(\mathcal{F}) \leq \mathcal{R}_N(\mathcal{F}) + (b-a)\left(\frac{\ln(2/\epsilon)}{\beta_2 N}\right)^{\frac{1}{\gamma}}.$$

However, the behavior of Rademacher complexity $\mathcal{R}(\mathcal{F})$ cannot be totally described by its empirical version $\mathcal{R}_N(\mathcal{F})$. In fact, the joint distribution of $\sum \sigma_n f(\mathbf{z}_n)$ may not be a Rademacher process because $\mathbf{z}$ is a random variable of an unknown distribution.

### 5.2 Answer to Question Q2

Recalling (10), the lower and the upper bounds of $\mathcal{R}_N(\mathcal{F})$ are respectively derived from the Sudakov minoration for Rademacher processes and Massart's finite class lemma, both of which are strongly related to (or have the similar forms as that of) the Hoeffding-type results.[7] Thus, all Hoeffding-type conclusions mentioned in Section 2.3 are consistent.

However, as answered above, the joint distribution of $\sum \sigma_n f(\mathbf{z}_n)$ may not be a Rademacher process. Since Sudakov minoration is not valid for an arbitrary process [see Talagrand, 1994b,a, Latała, 1997], the lower bound of $\mathcal{R}_N(\mathcal{F})$ with the rate $O(N^{-\frac{1}{2}})$ is not applicable to the lower bound of $\mathcal{R}(\mathcal{F})$.

Additionally, we also need to consider the following question: why we do not use Bennett's inequality to obtain the upper bound of $\mathcal{R}_N(\mathcal{F})$? In fact, in order to obtain the upper bound of $\mathcal{R}_N(\mathcal{F})$, one needs to consider the following term $\mathrm{E}\mathrm{e}^{\sigma f(\mathbf{z})}$. Since $\sigma$ is a Rademacher variable taking values from $\{\pm 1\}$ with equivalent probability for a given sample $\mathbf{z}$, there holds that

$$\mathrm{E}\mathrm{e}^{\sigma f(\mathbf{z})} = \frac{\mathrm{e}^{f(\mathbf{z})} + \mathrm{e}^{-f(\mathbf{z})}}{2} \leq \mathrm{e}^{\frac{(f(\mathbf{z}))^2}{2}},$$

which differs from the related formula (28) in Lemma A.1 that is built in the case that the distribution of $\mathbf{z}$ is unknown.

---

[7] Recalling Massart's finite class lemma, the main step of its proof is processed by using a term $\mathrm{e}^{\lambda^2 r^2/2}$ which is also similar to the term $\mathrm{e}^{\lambda^2 r^2/8}$ appearing in Hoeffding's lemma.
See http://ttic.uchicago.edu/~tewari/lectures/lecture10.pdf

## 6 Asymptotical Convergence

Based on the generalization bound (16), we study the asymptotic convergence of the i.i.d. learning process. We also give a comparison with the existing results.

Recalling (13), it is noteworthy that there is only one solution $x = 0$ to the equation $\Gamma(x) = 0$ and $\Gamma(x)$ is monotonically decreasing when $x \geq 0$ [see Fig. 3]. Following Theorem 4.1, we can directly obtain the following result indicating that the asymptotic convergence of the i.i.d. learning process is determined by the uniform entropy number $\ln \mathcal{N}_1(\mathcal{F}, \xi/8, 2N)$.

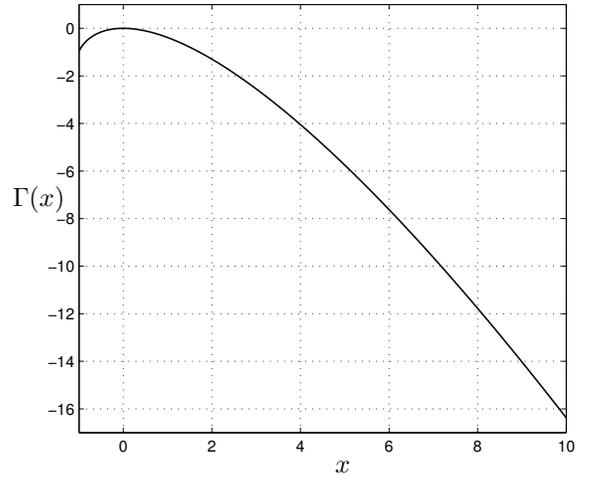

Figure 3: The Function Curve of $\Gamma(x)$

**Theorem 6.1** *Under the notations and conditions of Theorem 4.1, if the following condition is supported:*

$$\lim_{N \to +\infty} \frac{\ln \mathcal{N}_1(\mathcal{F}, \xi/8, 2N)}{N} < +\infty, \quad (24)$$

*then we have for any $\xi > 0$,*

$$\lim_{N \to +\infty} \Pr\left\{\sup_{f \in \mathcal{F}} \left|\mathrm{E}f - \mathrm{E}_N f\right| > \xi\right\} = 0. \quad (25)$$

As shown in Theorem 6.1, if the uniform entropy number $\ln \mathcal{N}_1(\mathcal{F}, \xi/8, 2N)$ satisfies the condition (24), the probability of the event

$$\sup_{f \in \mathcal{F}} \left| \mathrm{E}f - \mathrm{E}_N f \right| > \xi$$

will converge to *zero* for any $\xi > 0$, when the sample number $N$ goes to *infinity*. This is in accordance with the classical result shown in (7): the probability of the event that $\sup_{f \in \mathcal{F}} \left| \mathrm{E}f - \mathrm{E}_N f \right| > \xi$ will converge to *zero* for any $\xi > 0$, if the uniform entropy number $\ln \mathcal{N}_1(\mathcal{F}, \xi/8, 2N)$ satisfies the same condition as (24).

## 7 Conclusion

In this paper, we present the Bennett-type generalization bounds and analyze the rate of convergence of the derived bounds. In particular, we first extend the classical Bennett's inequality to develop two Bennett-type deviation inequalities. Based on the derived inequality, we then obtain the generalization bounds based on the uniform entropy number and the Rademcaher complexity, respectively.

Moreover, we show that the Bennett-type generalization bounds have a faster rate $o(N^{-\frac{1}{2}})$ of convergence than the rate $O(N^{-\frac{1}{2}})$ of the classical Hoeffding-type results [see (7) and (8)]. Especially, we show that the rate will become faster in the large-deviation case, where the empirical risk $\mathrm{E}_N f$ is far away from the expected risk $\mathrm{E}f$. In contrast, the Hoeffding-type results provide the rate $O(N^{-\frac{1}{2}})$ regardless of the discrepancy between $\mathrm{E}_N f$ and $\mathrm{E}f$. From view of this point, the Bennett-type results give a more detailed description to the asymmetrical behavior of the learning process.

We give an explanation to the "contradiction" mentioned in Remark 5.1. As shown in Talagrand [1994b,a], Latała [1997], Sudakov minoration provides a lower bound of the empirical Rademacher complexity $\mathcal{R}_N(\mathcal{F})$ with the rate $O(N^{-\frac{1}{2}})$ [Mendelson, 2003], because the empirical Rademacher complexity $\mathcal{R}_N(\mathcal{F})$ actually is a special case of subgaussian process given a sample set $\{\mathbf{z}_n\}_{n=1}^N$. In contrast, it is difficult to specify the distribution characteristics of $\mathcal{R}(\mathcal{F})$ and thus the classical result (10) could not be applied to bound $\mathcal{R}(\mathcal{F})$.